\documentclass[10pt,a4paper,conference]{IEEEtran}

\usepackage{cite}

\usepackage{hyperref}
\hypersetup{
    colorlinks=true,
    linkcolor=blue,
    filecolor=blue,      
    urlcolor=magenta,
}
\usepackage{multirow}
\ifCLASSINFOpdf
  \usepackage[pdftex]{graphicx}
  \graphicspath{{../pdf/}{../jpeg/}{./images/}}

  \DeclareGraphicsExtensions{.pdf,.jpeg,.png}
\else
  \usepackage[dvips]{graphicx}
  \graphicspath{{../eps/}}
  \DeclareGraphicsExtensions{.eps}
\fi

\usepackage{ulem}
\usepackage{subcaption}
\usepackage{amsmath}
\usepackage{array}
\usepackage{url}
\usepackage{adjustbox,lipsum, xcolor}
\usepackage{soul}

\usepackage{color, colortbl}
\usepackage[first=0,last=9]{lcg}

\definecolor{Gray}{gray}{0.9}

\def\SPSB#1#2{\rlap{\textsuperscript{#1}}\textsubscript{#2}}
\def\SP#1{\textsuperscript{#1}}
\def\SB#1{\textsubscript{#1}}

\newcommand\vgg{VGG}
\newcommand\resnet{ResNet50}
\newcommand\jpeg{JPEG}
\newcommand\coco{MS-COCO}
\newcommand\pascal{Pascal VOC}
\newcommand\imagenet{ImageNet}

\begin{document}
%
\title{Object Detection in the DCT Domain: is Luminance the Solution?}

\author{\IEEEauthorblockN{Benjamin Deguerre}
\IEEEauthorblockA{INSA Rouen Normandie\\
ACTEMIUM Paris Transport\\
Email: benjamin.deguerre@insa-rouen.fr}
\and
\IEEEauthorblockN{Clement Chatelain}
\IEEEauthorblockA{INSA Rouen Normandie\\
Email: clement.chatelain@insa-rouen.fr}
\and
\IEEEauthorblockN{Gilles Gasso}
\IEEEauthorblockA{INSA Rouen Normandie\\
Email: gilles.gasso@insa-rouen.fr}}


%

\maketitle

\begin{abstract}

Object detection in images has reached unprecedented performances. The state-of-the-art methods rely on deep architectures that extract salient features and predict bounding boxes enclosing the objects of interest. These methods essentially run on RGB images. However, the RGB images are often compressed by the acquisition devices for storage purpose and transfer efficiency. Hence, their decompression is required for object detectors. To gain in efficiency, this paper proposes to take advantage of the compressed representation of images to carry out object detection usable in constrained resources conditions.
 
Specifically, we focus on \jpeg{} images and propose a thorough analysis of detection architectures newly designed in regard of the peculiarities of the \jpeg{} norm. This leads to a $\times 1.7$ speed up in comparison with a standard RGB-based architecture, while only reducing the detection performance by 5.5\%. Additionally, our empirical findings demonstrate that only part of the compressed \jpeg{} information, namely the luminance component, may be required to match detection accuracy of the full input methods. Code is made available at : \url{https://github.com/D3lt4lph4/jpeg_deep}.

\end{abstract}

%
\IEEEpeerreviewmaketitle

\section{Introduction}

Deep architectures, especially Convolutional Neural Networks (CNN) have become a standard for object detection. Most of the proposed architectures rely on the same two components: a feature extractor (backbone) pre-trained on a classification task and a detection head to output the box predictions.  Impressive results were achieved on datasets growing in complexity such as \pascal{} \cite{pascalvoc} or \coco{} \cite{mscoco}. However, even though most of the images are compressed in order to limit the storage and the transfer bandwidth requirements,  state-of-the-art detection architectures are designed for processing RGB inputs. Hence the usual procedure for any detection task observes the two following steps:
\begin{itemize}
    \item The image is uncompressed and possibly pre-processed, namely with normalisation and/or resizing operations.
    \item Then the detection task is performed, typically using a deep network.
\end{itemize}

Although effective, this procedure has some drawbacks as the image decoding step induces a computational cost. Moreover, the involved deep architectures  may include a tremendous amount of parameters, due to the RGB image resolution, making the network computationally expensive. These facts  hinder the large scale deployment of these networks for applications with real-time and computation constraints such as city surveillance \cite{citywide_traffic_estimation} or road traffic monitoring and management \cite{comp_count}.

In this article, we propose to directly perform object detection in compressed \jpeg{} images, \jpeg{} being one of the most popular and effective compression algorithms. 
As the \jpeg{} compression transform images in a tiled frequency space, through Discrete Cosine Transform (DCT), it raises the question whether a detection network is able to efficiently map the frequency domain into a spatial domain in order to output the position of the objects in the original image.
Detection in a \jpeg{} compressed domain is also faced to the chrominance ($C_r$ and $C_b$ component) sub-sampling that is usually performed, that may lead to a resolution change when compared to the luminance component ($Y$ component).

Our preliminary work \cite{dct_detection}  demonstrates that object detection in \jpeg{} images is feasible.
By taking advantage of the DCT blocks representation of the image to reduce the overall computation cost, we were able to speed up the inference stage by a factor of 2, at the cost of a reduced accuracy, 45\% lower than the detection performances of the RGB-based architectures.
Furthermore, the aforementioned sub-sampled chrominance problem was not addressed.

The present work shows that not only object detection in the compressed domain is achievable, but also that very close detection performance to those of RGB-based architectures can be reached, still with a significant speed up gain.
Additionally, we investigate the use of the sole $Y$ channel as input of the proposed detection networks, the benefit being the reduction of the required bandwidth.
We empirically demonstrate that using only the $Y$ channel is enough to reach an accuracy equivalent to the one of networks relying on $YC_bC_r$.

To reach these goals, we train several backbone classification networks designed to account for the specificity of the \jpeg{} norm.
With regard to the detection task, we incorporate these networks within a SSD \cite{ssd}, and evaluate the overall designed detector on two common detection benchmarks, \pascal{} \cite{pascalvoc} and \coco{} \cite{mscoco}.
Finally, as a side contribution, we provide a thorough analysis of the learned classification networks, in view of usage on restricted computational resources.
We evaluate the classification networks on the \imagenet{} \cite{imagenet_dataset} dataset and show that networks that may be highly efficient when evaluated with powerful GPUs may behave poorly in more constrained settings.

To summarise, our contributions are threefold:

\begin{itemize}
    \item We show that detection in the compressed domain can nearly reach detection performance of the RGB domain, and highlight its interest in embedded conditions. This is achieved by embedding adapted classification networks in a detection architecture such as the SDD \cite{ssd}.
    \item We provide a thorough evaluation of several detection models (backbone + SSD) in the compressed domain on two public datasets, namely \pascal{} and \coco{}.
    \item And finally, we experimentally show that using only the luminance is enough for detection, effectively reducing the need for bandwidth.
\end{itemize}

The rest of the paper is as follows: Section \ref{r_work} goes over the existing literature and Section \ref{jpeg_norm} presents the JPEG norm. The proposed approach and architectures are detailed in Section \ref{p_a}. Section \ref{results} present the obtained results and discusses on the strength and limitation of compressed JPEG inputs. 

\section{Related works}
\label{r_work}

This article focuses on object detection in embedded conditions. Section \ref{obj_detec} gives related works to object detection in RGB images, while section \ref{deep_comp} details existing literature on the usage of the compressed representation of data coupled with deep learning. 

\subsection{Object detection}
\label{obj_detec}

Object detection   aims at detecting multiple objects within an image.
This is generally achieved by finding the coordinates of bounding boxes around objects in the images.
Methods for object detection are broadly based on RBG inputs and rely on two main deep architectures: two-stage detectors and one-stage detectors.
The former networks are based on a region proposal stage followed by a box classification stage while the latter architectures use anchor-boxes and predict a set of pre-defined boxes as object or background.

Two-stage detectors were introduced with R-CNN \cite{rcnn}, using Selective Search \cite{s_search} as region proposal. Further, R-CNN was improved in Fast R-CNN \cite{fast_rcnn} with feature computation sharing and in Faster R-CNN \cite{faster_rcnn} with a Region Proposal Network. Optimising object detection and object segmentation simultaneously, Mask R-CNN \cite{mask_rcnn} yielded another level of performance improvement.  Trying to reduce the number of close multiple detection of the same object, Cascade R-CNN \cite{cascade_rcnn}, proposes a cascaded detection pipeline using increasing Intersection over Union (IoU) thresholds to select the positive examples used for training. CBNet \cite{CBNet}, the current top-1 ranking approach in the \coco{} detection challenge \cite{mscoco} is built on Mask R-CNN and uses cascaded classification backbone networks to improve the accuracy of the detector.

One-stage detectors were popularised by YOLO \cite{yolo} and SSD \cite{ssd}. While originally suffering  a lower accuracy than SSD, YOLO was successively improved \cite{yolov2,yolov3} through anchor-boxes K-means selection, multi-label object class prediction or prediction across scales.  Retina\-Net \cite{retinanet} proposed the focal loss to alleviate the background imbalance problem for one-shot detectors. Using a unified module prediction and connection between low and high prediction layers, \cite{res_co_uni_module} further improved the accuracy of single stage detectors. Rep\-Points \cite{RepPoints}, gets rid of the rectangular anchor boxes through point prediction. Using a bi-directional feature pyramid network to share information between the different scales, Efficient\-Det is currently the best performing one-stage detector on the \coco{} dataset, behind CBNet \cite{CBNet}.

While two-stage detectors tend to be more accurate, one-stage detectors provide faster inference time. Because we focus on settings with limited computational resource, we opt for one-stage detectors. 
Of particular note, 
many methods were proposed  to speed up the inference stage of RGB-based architectures without harming the detection performances
\cite{tinyssd} \cite{mobilenet} \cite{efficientdet}. The approach we propose rather takes another viewpoint as it aims at improving the network's inference speed by considering light-weight input $YC_bC_r$ or $Y$. 
Although out of the scope of this paper, the combination of our approach with the aforementioned methods could be of interest.

\subsection{Deep-Learning on compressed images/videos}
\label{deep_comp}

Using compressed images has been explored in the past for various computer vision tasks. Many applications estimating flows from videos take advantage of the compression format as the encoded data often include displacement information in order to reduce the size of the videos by exploiting temporal redundancy. Wang \textit{et al.}  \cite{fast_od_comp_vid} proposed an architecture for object detection within compressed videos. They use motion vectors and residuals to infer objects through time, while only partially decoding the compressed video flux.  Wu \textit{et al.} \cite{comp_ac_rec} also proposed to exploit encoded motion vectors and residuals to improve both accuracy and inference speed for a task of action recognition in videos. Taking another approach, \cite{dmc-net} detects action by first generating an optical flow from motion vectors and residuals and then using it to classify the action. We shall mention that, while not coupled with deep learning, motion vectors and/or DCT coded frames were leveraged on to count vehicles \cite{comp_count} or to estimate the vehicles' speed and density on highways \cite{highway_mpeg}.

Besides videos, compressed representations were considered for image classification, segmentation or detection tasks. Gueguen \textit{et al.} \cite{dct_classification} investigated different network architectures for JPEG image classification. They  reached state-of-the-art classification performance while speeding-up the prediction pass, even for their architectures requiring more FLOPs than their RGB counterparts. The gain for such architectures mainly stem from the reduced data transfer between CPU and GPU due to the image compression.
In \cite{dct_segmentation}, image segmentation using JPEG compressed images is proposed. The methods  adapts a RGB network by removing the down-sampling blocks to match the shape of the DCT inputs. The authors show impressive results almost reaching the RGB baseline with similar level of FLOPs. Finally, relying on JPEG2000 compression norm, Lahiru et al. \cite{jpeg2000_classification} showed results matching the RGB baseline as well as speed improvements for classification. To do so, they stack sub bands of half decoded images and feed them to a modified neural network.

\section{JPEG Norm}
\label{jpeg_norm}

JPEG encoding is a lossy compression algorithm. It relies on the human eye sensibility to chroma components in an image. JPEG norm exploits the sparsity of the DCT (Discrete Cosine Transform) representation of an image to carry the compression. 

\begin{figure}[!ht]
    \centering
    \includegraphics[width=\columnwidth]{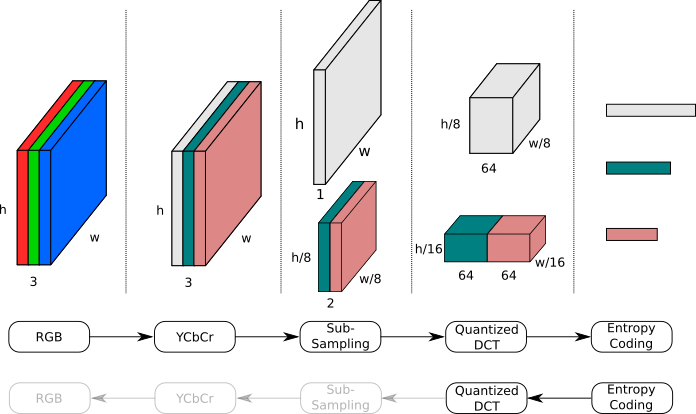}
    \caption{JPEG CODEC pipeline. Top: obtained representation through the encoding steps, Middle: encoding steps, Bottom: decoding steps; the faded boxes show the unused steps. $Y$ is represented in gray, $C_b$ in dark green and $C_r$ in faded pink. Image adapted from \cite{dct_classification}.}
    \label{fig:JPEG_pipeline}
\end{figure}

The whole JPEG compression/decompression pipeline is described in figure \ref{fig:JPEG_pipeline}.
The compression procedure is as follows: first the RGB image is converted to the $YC_bC_r$ domain. The human eye being less sensitive to the $C_b$ and $C_r$ components, this allows for easy compression by subsampling (usually $1/2$). Then a blockwise DCT (of size 8x8 pixels) followed by a quantization is applied. The compressed image is attained using the RLE/Huffman entropy coding. 

As shown in figure \ref{fig:JPEG_pipeline}, up until the entropy coding, the processed data is similar to an image in shape, and, for two images of the same size, the compressed representation will have the same size. Because quantization and Huffman compression are subject to variations depending on the compression level, we choose to use the de-quantized DCT coefficients as input to the models we propose (see figure \ref{fig:JPEG_pipeline}, bottom).

\section{Proposed approach}
\label{p_a}

\subsection{Detection in the frequency domain}

\begin{figure}
    \centering
    \includegraphics[width=\linewidth]{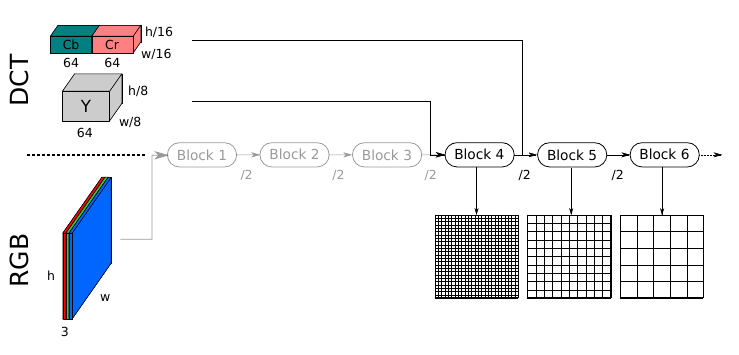}
    \caption{Principle of the DCT-based object detector. The grids at the bottom right represent the boxes predicted by the network. The further we advance in the network, the bigger the predicted boxes. Depending on the setup, $YC_bC_r$ or only $Y$ inputs are fed to the networks.
    For clarity, not all the prediction layers are shown.
    }
    \label{fig:detection}
\end{figure}

Our goal is to design an object detection network starting from compressed JPEG images. The main challenge lies in classifying and localising an object using frequency domain information, namely the DCT block coefficients of the intended image. Indeed, contrary to RGB images, the spatial information relative to the objects may be impeded by the DCT transform. We lay our proposal on SSD \cite{ssd} which was originally designed to work on RGB inputs. SSD is based on a multi-scale box prediction, each of the successive feature maps are designed to predict boxes of different sizes. In order for the SSD to work with JPEG compressed inputs, we need to modify it. As shown in Figure \ref{fig:detection}, due to the blockwise DCT, $Y$ inputs to the network are downsized by 8 and the number of channel is augmented to 64 when compared with the original RGB image. Furthermore, due to the down-sampling on the chrominance components, we can see the $C_b$ and $C_r$ inputs further downsized by a factor of two. Hence, the first blocks of the genuine SSD are skipped (or replaced by blocks without down-sampling). 

\subsection{Design of the backbones, \texorpdfstring{$YC_bC_r$}{YCbCr} input and \texorpdfstring{$Y$}{Y} input}

As previously mentioned,  the \jpeg{} compression algorithm requires to design carefully the detection network. Indeed, the $C_b$ and $C_r$ components are down-sampled, which leads to input matrices smaller by a factor of two for these components. As a result, the smallest predicted boxes are not provided with the chromatic information. This is illustrated in Figure \ref{fig:detection}. To fix this, we propose to use a deconvolution layer to scale up the $C_b$ and $C_r$ components to use them in combination with the $Y$ component.

Because of the change in resolution induced by the JPEG compression algorithm, we make the hypothesis that all the components may not be useful. Intuitively, as the sub-sampling operation is carried before the blockwise DCT, the $C_b$ and $C_r$ $8\times8$ blocks are actually a sparse representation of an equivalent four $Y$ $8\times8$ blocks. Hence, the learning algorithm has to deal with both sparsity and resolution problems when the $C_b$ and $C_r$ inputs are concatenated with the main part of the networks, potentially impeding the training. We thus propose to test a variation of the networks that only uses the $Y$ component as input.

\subsection{Proposed architectures}
\label{subsec:proposed_approach}
We now detail all of the tested architectures. Originally, the SSD was built on top of a VGG classification network. Besides modified VGGs for frequency domain, we also investigate \resnet{} based classification backbones as such networks have provided impressive results in terms of accuracy and inference speed for JPEG image classification \cite{dct_classification}. The features of these detection architectures are  summarised in table \ref{implementations}. Notice that the networks  using the sole $Y$ channel are not detailed as they are simplified instances of the $YC_bC_r$-based networks.

We start by reviewing the RGB architectures in \ref{rbg_baselines}. Then in \ref{ycbcr} we detail the architectures using the compressed inputs that do not use deconvolution layers. And we finish with the deconvolution architectures in \ref{ycbcr_deconv}.

\begin{table*}[!ht]
    \caption{Features of the proposed networks. A * indicates that the layer is used for boxes prediction. The lines are arbitrary and do not represent the shape of the layers. Note that LC-RFA-thinner is skipped  as it is  a variation of the LC-RFA model.
    }
    \centering
    \begin{subtable}{1\textwidth}
    \begin{adjustbox}{max width=\linewidth}
    \begin{tabular}{ l l l l l l l l l l}
        \hline
            SSD \cite{ssd} & \multicolumn{2}{c}{SSD DCT \cite{dct_detection}} & \multicolumn{2}{c}{SSD DCT-deconv} & SSD ResNet & \multicolumn{2}{c}{SSD LC-RFA} & \multicolumn{2}{c}{SSD Deconvolution-RFA} \\
             & \multicolumn{2}{c}{(reference: SSD)} & \multicolumn{2}{c}{(reference: SSD)} & (reference: SSD) & \multicolumn{2}{c}{(reference: SSD ResNet)} & \multicolumn{2}{c}{(reference SSD ResNet)} \\
        \hline
        
            RGB \emph{(300,300,3)} & Y \emph{(38,38,64)} & Cb, Cr \emph{(19,19,128)} & Y \emph{(38,38,64)} & Cb, Cr \emph{(19,19,128)} & RGB \emph{(300,300,3)} & Y \emph{(38,38,64)} & Cb, Cr \emph{(19,19,128)} & Y \emph{(38,38,64)} & Cb, Cr \emph{(19,19,128)} \\

            \rowcolor{Gray}
            C\SB{11}, C\SB{12} &  &  &  &  &  &  &  &  & \\
            \rowcolor{Gray}
            P\SB{1} & & & & & C(64,7,2) & & & & \\

            C\SB{21}, C\SB{22} &  &  &  &  & BN, R &  &  &  & Deconv(28, 28, 128)\\
            P\SB{2} &  &  &  &  & M(3,2) &  &  & Concat(39,39,192) & - - - -\\

            \rowcolor{Gray}
            C\SB{31}, C\SB{32}, C\SB{33} &  &  &  & Deconv(38,38,128) & CB\SB{2}(s=1) & BN,CB\SB{4}(k=1,s=1) &   & CB\SB{4}(k=1,s=1) & \\
            \rowcolor{Gray}
            P\SB{3} & BN, C(256,3,1) & & Concat(38,38,192) & - - - - & IB, IB & IB(k=2), IB &  & IB(k=2), IB & \\
            
            C\SB{41}, C\SB{42}, C\SPSB{*}{43} & C\SB{41}, C\SB{42}, C\SPSB{*}{43}  &  &  BN, C\SB{41}, C\SB{42}, C\SPSB{*}{43} &  & CB\SB{3} & CB\SB{3} &  & $\leftarrow$ & \\
            P\SB{4} & P\SB{4} & BN & P\SB{4} & & IB, IB, IB\SP{*} & IB, IB, IB\SP{*} & & &\\
             & Concat & - - - - & & & & CB\SB{4} & BN, CB\SB{4}(k=1, s=1) & & \\

            \rowcolor{Gray}
            C\SB{51}, C\SB{52}, C\SB{53}   & $\leftarrow$ &  & $\leftarrow$  &  & CB\SB{4} & Concat & - - - - & & \\
            \rowcolor{Gray}
            P\SB{5} & & & & &  IB, IB, IB, IB, IB & $\leftarrow$  & & & \\
            \rowcolor{Gray}
            fc6, fc7 & & & & & CB\SB{5}(s=1), IB, IB, C\SB{61} & & & & \\
            \rowcolor{Gray}
            C\SB{61}, C\SPSB{*}{62} & & & & & C\SPSB{*}{62} &  &  &  &\\            
            
            C\SB{71} &  &  &  &  & $\leftarrow$ & & & & \\
            C\SPSB{*}{72} & & & & & &  &  &  &\\

            \rowcolor{Gray}
            C\SB{81}, C\SPSB{*}{82} &  &  &  &  &  & & & &  \\

            C\SB{91}, C\SPSB{*}{92} &  &  &  &  & & & & & \\
        \end{tabular}
    \end{adjustbox}
    \end{subtable}
    
    \bigskip
    \begin{subtable}{0.5\textwidth}
    \centering
    \begin{adjustbox}{max width=\linewidth}
    \begin{tabular}{| l p{5cm} l p{5cm} |}
        \hline
            \textbf{Legend} &  &  &  \\
        
            RGB & RGB pixel input  & Deconv & Deconvolution with 64 output channels, filter size 2, stride 2. Separate deconvolution layers are applied to Cb and to Cr, resulting in 128 total output channels \\
            Y & Y channel DCT input & BN & BatchNormalization \\
            Cb, Cr & Cb and Cr channel DCT input & R & Relu \\
            C\SB{ij} & J-th convolution layer of the I-th block in the original SSD architecture & CB\SB{n} & ConvBlock stage n, with number of channels as in original ResNet-50 paper, kernel size = 3 and stride = 2 unless specified otherwise. \\
            P\SB{i} & Pooling layer of the I-th block in the original SSD architecture & IB & IdentityBlock, with number of channels matched to preceding CB layer (as in ResNet-50) \\
            
            C & Convolution(channels, filter size, stride) & M & MaxPooling(pool size, stride) \\
            Concat & Channel wise concatenation & $\leftarrow$ & Layers after this point are the same as reference \\
            
            - - - - &  Channel is being concatenated & * & Layer is used for boxes prediction \\
        \hline 
        \end{tabular}
    \end{adjustbox}
    \end{subtable}

    \label{implementations}
\end{table*}

\subsubsection{RGB baselines}
\label{rbg_baselines}

For each of the backbones, we compare our results with the RGB based architecture. For the \vgg{} based SSD, we use the original architecture  \cite{ssd}. In order to use the \resnet{} as backbone, few modifications must be applied to the original SSD. Originally, part of the SSD convolutional layers, the fc6 and fc7 layers (c.f Table \ref{implementations}), were designed to use the \vgg{} dense layers' weights. As the \resnet{} does not contains such layers, we need to modify the SSD architecture to correctly incorporate the \resnet{} backbone. We remove all the original SSD layers up to (including) the fc7 layer and replace them with the \resnet{} ones (except for the last classification layer). This way, as for the \vgg{}, only the weights from the classification layer are not pre-loaded into the SSD.

\subsubsection{\texorpdfstring{$YC_bC_r$}{YCbCr} DCT methods}
\label{ycbcr}

We present the DCT architectures that do not account for the discrepancies between the $Y$ and $C_b$, $C_r$ components.
Hence, for all the architectures of this subsection, the smallest boxes only rely on the $Y$ input for predictions (c.f Figure \ref{fig:detection}).
The \vgg{}-based architecture, SSD DCT, is directly re-implemented from \cite{dct_detection}. For the \resnet{}-based architectures, we select two classification networks, Late-Concat-RFA (LC-RFA) and Late-Concat-RFA-Thinner, which have shown good precision/accuracy ratios as evidenced in \cite{dct_classification}.
RFA stands for Receptive Field Aware.
The LC-RFA architectures imitate the original \resnet{} receptive field by removing the downsizing carried by some of the \resnet{} convolution layers.
LC-RFA-Thinner is a lighter, hence faster, variant of LC-RFA.

\paragraph{SSD DCT}
\label{par:ssd_dct}

This method is based on the original SSD, the first three convolution blocks are removed and the $Y$ and $C_b$,$C_r$ inputs are respectively plugged in the fourth and fifth block. Due to the hard pruning of the first blocks, this is one of the fastest detection architecture proposed.

\paragraph{SSD LC-RFA}
\label{par:ssd_lcrfa}

We integrate the LC-RFA classification network in the same way as the \resnet{} SSD RGB version. This architecture does not prune away the first convolution blocks, instead, the downsizing operations are removed by changing the stride of the convolutions from 2 to 1 when required. It is worth noticing that we had to modify part the original architecture from \cite{dct_classification} as the one given by the authors lead to channel incompatibility on some of the layers. Specifically, we replace the last two CB\SB{4} blocks along the $Y$ and $C_b$, $C_r$ axis with CB\SB{3} blocks. 

\paragraph{SSD LC-RFA-Thinner}
\label{par:ssd_lcrfat}
It is a lighter version of SSD LC-RFA with a reduced number of layers \cite{dct_classification}.
 This approach increases detection speed while keeping fairly good accuracy. Modifications in the number of channels is as follows: along the $Y$ axis, channels from the three first CB blocks are set from \{1024, 512, 512\} to \{384, 384, 768\} and along the $C_b$, $C_r$ axis, the channels of the CB block set from \{512\} to \{256\}.

\subsubsection{$YC_bC_r$ DCT Deconvolution methods}
\label{ycbcr_deconv}
We use deconvolution methods to align the size of the down-sampled $C_b$, $C_r$ to the one of $Y$. The related architectures are given below. 

\paragraph{SSD DCT-Deconv}
\label{par:ssd_dct_deconv}

It is based on a VGG where the first three blocks are skipped. The $C_b$ and $C_r$ inputs first go through a deconvolution layer each and are concatenated with the $Y$ component. Then a Batch Normalization is applied and outputs the input of the fourth block. The rest of the network follows the original SSD.

\paragraph{SSD Deconvolution-RFA}
\label{par:ssd_decovolution_rfa}

It is based on \resnet{}. We test using the deconvolution module proposed in \cite{dct_classification}. Contrary to SSD DCT-Deconv, the first blocks are not skipped, instead the stride of the first convolutions is changed from 2 to 1 when required. This architecture is mostly equivalent in speed and accuracy to the LC-RFA network for classification purpose.

\section{Experiments and results}
\label{results}

Experiments were conducted to evaluate the investigated detection networks. As a preliminary, we first implement and train the classification networks  \resnet{}, VGG, LC-RFA and their variants (see Section \ref{subsec:proposed_approach}) using JPEG images (namely their block DCT parameters). For this, we use the Imagenet2012 training set \cite{ILSVRC15}. The learned DCT-based networks are then evaluated on the validation set. 

As for detection, we train and evaluate on two datasets, \pascal{} and \coco{}. All reported results on \pascal{} are the models evaluation on 2007 test set. For the \coco{} dataset, we use the standard train-validation-test sets.

\subsection{Implementation details}

\paragraph{Classification}
\label{par:classification_imp}

We train the classification networks using a distributed environment and follow the recommendations from \cite{distributed_training}. The models were trained using horovod on 4 nodes amounting to a total of 8 GPUs. Default training parameters were used for VGG and ResNet as described in \cite{vgg, resnet_le_bon}. For data-augmentation, we rescale the images so that the smallest side is 256 pixels, we randomly crop a 224x224 patch and then randomly apply horizontal flip. The learning rate is decayed by 10 whenever the validation loss plateaued. While the original articles apply weight decay on the loss, due to framework limitation, we use a per layer one. We found out later that in this setting, the weight decay should actually be reduced by a factor of 2. Given that we get results close to the authors' baselines for the RGB networks, we keep the setting. The trained classification networks served as backbone for the detection networks. 

We evaluate the number of Frames Per Seconds (FPS) that can be processed for each of the networks. We use a NVIDIA GTX 1080, set the batch size to 8 and do 10 runs of 200 predictions. The final FPS value is the average over the runs. FPS from other papers (when provided by the authors) are are not directly reported as they used different GPU implementations. 

\paragraph{Detection}
The SSD-based detection networks were trained on a single GPU. The networks were initialised using the weights from the corresponding classification networks.

For the VGG based SSD, we follow \cite{ssd} and convert the dense classification layers into convolutional layers. When converting the VGG's dense layers weights to fit the convolution layers from the SSD, we use a pre-set sub-sampling of 0:4:4096 to extract 1024 channels from the original 4096. 

For the PASCAL VOC, we train on two different sets, the original 2007 training/validation and the 2007+2012 training/validation. They are respectively denoted as 07 and 07+12 in the result tables. For the MS-COCO, we use the provided training/validation datasets.

We evaluate the FPS of each detection network using the same parameters as for classification. When evaluating the speed of the networks we find the Non-Maximum Suppression to be the limiting factor. For some of the architectures, this led to a sub-optimal usage of the GPU's capacities, especially for the VGG-DCT based networks. 
To account for this, we report two speed evaluations: i) with one model instantiated on the GPU, and ii) with two models instantiated on the GPU. We stopped at two instances as the models were saturating the capacities of the GPU.

\subsection{Evaluation of the classification networks}

\begin{figure}
    \centering
    \includegraphics[width=\linewidth]{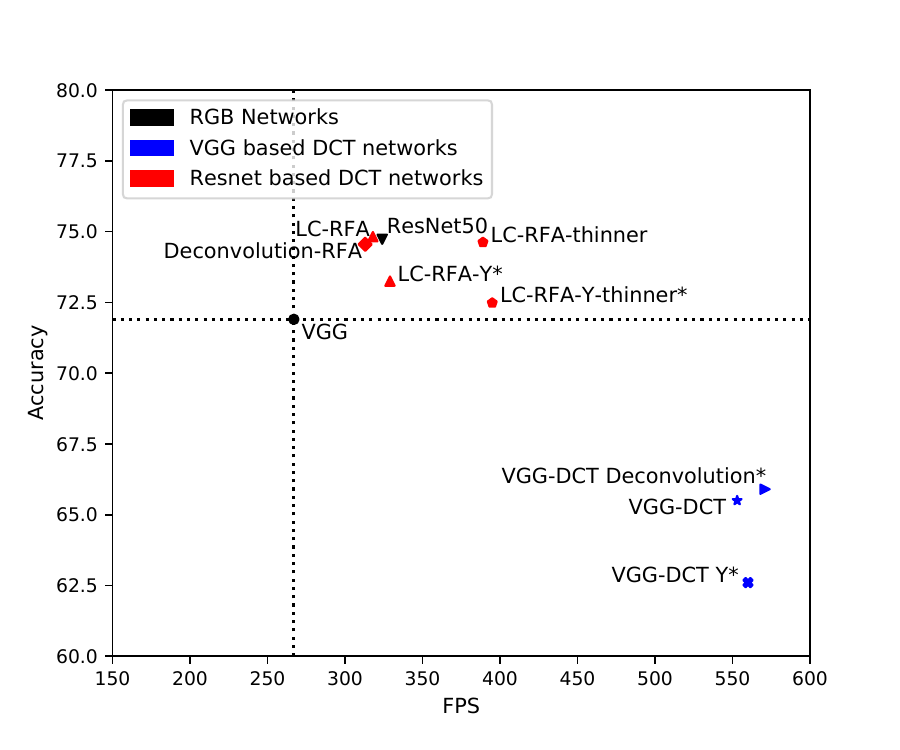}
    \caption{Accuracy vs FPS for the classification networks. The starred networks are the ones presented in this paper.}
    \label{fig:accuracy_fps_classification}
\end{figure}

We rescale the smallest size of the images to 256 and keep the proportions. We feed them to the networks and average the predictions through a Global Average Pooling layer whenever required. For each of the networks, we also retrain on RGB images to set a baseline given our data-augmentation. Results on the \imagenet{} validation dataset are reported in Table \ref{table:classification_ilsvrc_results} and the Accuracy vs FPS is shown in Figure \ref{fig:accuracy_fps_classification}.

\paragraph{Training with $YC_bC_r$ DCT inputs}

We first retrain all the architecture presented in \cite{dct_classification}, namely LC-RFA, LC-RFA-Thinner and Deconvolution-RFA,  and get accuracy results that are about 1$\sim$2 \% lower that the original ones. We attribute these differences to small changes in the evaluation method as well as some possible differences in hyper-parameters as they were not fully disclosed. The main difference concerns the FPS of the networks, we do not reproduce the same speed improvements between the RGB and DCT architectures. The main gains are obtained for the LC-RFA-Thinner architecture when compared with RGB with a $\times 1.2$ ($\times 1.77$ in \cite{dct_classification}) speed improvement, at equivalent accuracy. We believe these differences are due to the different GPU used for the evaluation. Our testing GPU does not process the images at a rate sufficient to take advantage of the reduce data transfer between CPU and GPU entailed by the compressed inputs. 

We retrain the VGG-DCT presented in \cite{dct_detection} with our training pipeline and show improvements of 41\% on the error (23.5 points) while showing the same speed-up at $\times 2.1$ (vs $\times 2$ for \cite{dct_classification}) when compared with RGB networks. The Deconvolution version of this network performs a bit better with an accuracy of 65.9 and improves the FPS by $\times 2.2$ when compared with RGB VGG.

If we compare the VGG based networks with the \resnet{} based ones, we can see that the VGG networks are about a 50\% faster but increase the error by 9 points (34\%). We attribute these differences to the hard pruning done on the first layers of the RGB architecture. 

\paragraph{Training with only $Y$ DCT input}

The related classification networks are respectively denoted as VGG-DCT Y, LC-RFA Y and LC-RFA-Thinner Y. The obtained results lead to the following remarks: the accuracy slightly decreases while the networks' speed increases. The smallest decrease is for the LC-RFA Y architecture with 3.1\% drop in accuracy while the biggest is for the VGG-DCT Y network with a decrease of 4.4\%. This seems to be consistent with the fact that \resnet{} classifier is more accurate than the VGG. The speed improvements,  ranging from 6 to 30 FPS,  are due to the reduction in computation entailed by only using the $Y$ input. While FPS gain may be negligible when comparing with the loss in accuracy, the reduced bandwidth due to the usage of the $Y$ component only makes such architecture attractive in case of limited computation resources. 

\begin{table}[!ht]
    \caption{Classification results on \imagenet{}. The top panel refers to RGB-based networks, the second one corresponds to DCT-based architectures. The two last panels refer to our implementations. In bold are the best results of our trainings.}
    \centering
    \begin{adjustbox}{max width=\linewidth}
        \begin{tabular}{l c c c }
        \hline
            Network & top-1 accuracy & top-5 accuracy & FPS \\
        \hline
            \emph{State of the Art:} & & & \\
            VGG \cite{vgg} & 73.0 & 91.2 & N/A \\
            VGG-DCT \cite{dct_detection} & 42.0 & 66.9 & N/A \\
        \hline
            Resnet50 \cite{dct_classification} & 75.78 & 92.65 & N/A \\
            LC-RFA (DCT) \cite{dct_classification} & 75.92 & 92.81 & N/A \\
            LC-RFA-thinner (DCT) \cite{dct_classification} & 75.39 & 92.57 & N/A \\
            Deconvolution-RFA (DCT) \cite{dct_classification} & 76.06 & 92.02 & N/A \\
        \hline
        \hline
            \emph{our trainings (VGG based):} & & & \\
            VGG & 71.9 & 90.8 & 267 \\
            VGG-DCT & 65.5 & 86.4 & 553 \\
            VGG-DCT Y & 62.6 & 84.6 & \textbf{583} \\
            VGG-DCT Deconvolution & 65.9 & 86.7 & 571 \\
        \hline
            \emph{our trainings (ResNet50 based):} & & & \\
            Resnet50 & 74.73 & 92.33 & 324 \\
            LC-RFA (DCT) & \textbf{74.82} & \textbf{92.58} & 318 \\
            LC-RFA Y (DCT) & 73.25 & 91.40 & 329 \\
            LC-RFA-Thinner (DCT) & 74.62 & 92.33 & 389 \\
            LC-RFA-Thinner Y (DCT) & 72.48 & 91.04 & 395 \\
            Deconvolution-RFA (DCT) & 74.55 & 92.39 & 313 \\
        \hline
            
        \hline
        \end{tabular}
    \end{adjustbox}
    \label{table:classification_ilsvrc_results}
\end{table}

{}

\subsection{Detection}

\begin{figure}
    \centering
    \includegraphics[width=\linewidth]{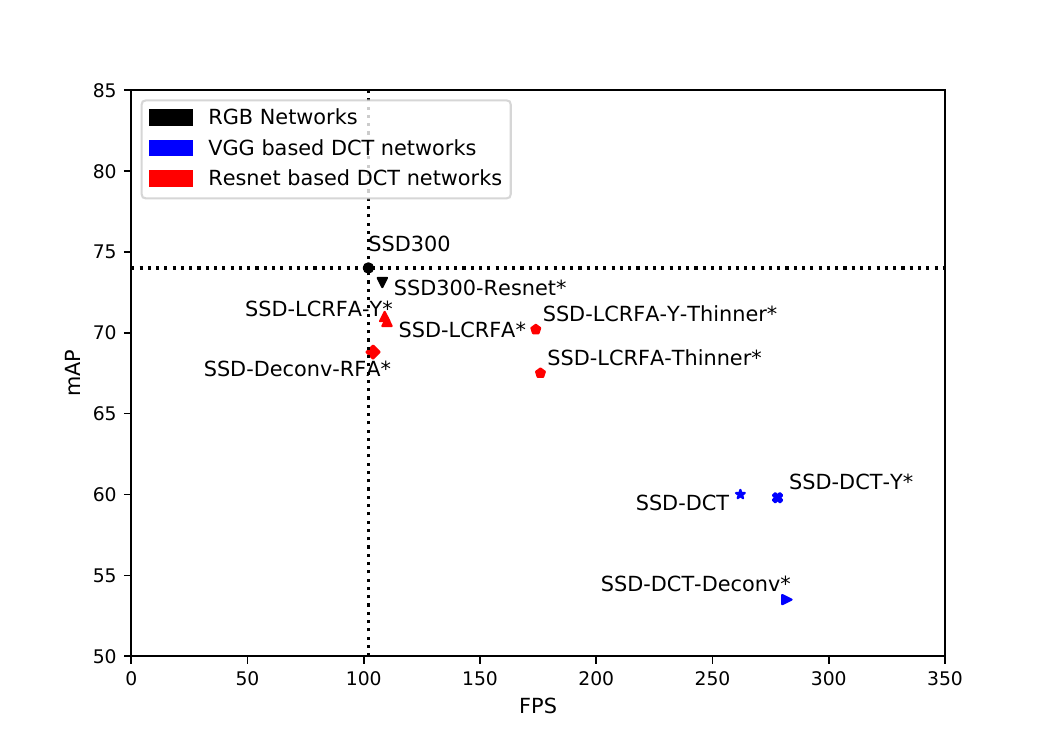}
    \caption{mAP vs FPS for the detection networks. Networks with a star at the end of their names are the one presented in this paper.}
    \label{fig:map_fps}
\end{figure}

We train the detection networks using the classification networks discussed previously. For fairness of comparison, we also retrain the RGB networks with our trained classification networks. The results for the PASCAL VOC evaluation are reported in table \ref{table:ssd-07+12} and the \coco{} results are reported in table \ref{table:ssd-coco}. We evaluate the speed of inference on the networks trained on the 07+12 PASCAL VOC dataset. Accuracy vs Speed is detailed in figure \ref{fig:map_fps} and the speed results are shown in table \ref{table:ssd-speed}.

\begin{table}[!ht]
    \caption{Detection results on the PASCAL VOC 2007 test set, 07 is for trained on 2007 data and 07+12 means trained on 2007+2012 data. The last two panels report the performances of our trained networks. In bold are the best results of our trainings.}
    \centering
    \begin{adjustbox}{max width=\linewidth}
        \begin{tabular}{l c c c}
        \hline
            Network & mAP (07) & mAP (07+12) & FPS \\
        \hline
            \emph{SoT:} & & & \\
            SSD300 \cite{ssd} & 68.0 & 74.3 & N/A\\
            SSD300 DCT \cite{dct_detection} & 39.2 & 47.8 & N/A\\
        \hline
        \hline
            \emph{our trainings (VGG based):} & & & \\
            SSD300 & \textbf{65.0} & \textbf{74.0} & 102 \\
            SSD300 DCT & 48.9 & 60.0 & 262 \\
            SSD300 DCT Y & 50.7 & 59.8 & 278 \\
            SSD300 DCT Deconvolution & 38.4 & 53.5 & \textbf{282} \\
        \hline
            \emph{our trainings (ResNet50 based):} & & & \\
            SSD300-Resnet50 & 61.3 & 73.1 & 108 \\
            SSD300 DCT LC-RFA & 61.7 & 70.7 & 110 \\
            SSD300 DCT LC-RFA Y & 62.1 & 71.0 & 109 \\
            SSD300 DCT LC-RFA-Thinner & 58.5 & 67.5 & 176 \\
            SSD300 DCT LC-RFA-Thinner Y & 60.6 & 70.2 & 174 \\
            SSD300 DCT Deconvolution-RFA & 54.7 & 68.8 & 104 \\
        \hline
        \end{tabular}
    \end{adjustbox}
    \label{table:ssd-07+12}
\end{table}

\begin{table}[!ht]
    \caption{Speed inference of the tested detection networks. The tests were performed on a GTX 1080, "1 inst." means that only one instance of the model was loaded on the GPU for testing, "2 inst." means that two instances of the model were loaded on the GPU.}
    \centering
    \begin{adjustbox}{max width=\linewidth}
        \begin{tabular}{c l c c }
        \hline
            & Network &  FPS (1 inst.) & FPS (2 inst.) \\
        \hline
            \parbox[t]{2mm}{\multirow{4}{*}{\rotatebox[origin=c]{90}{VGG}}} & SSD300  &  88 & 102 \\
            & SSD300 DCT & 136 & 262 \\
            & SSD300 DCT Y & 140 & 278 \\
            & SSD300 DCT Deconvolution & \textbf{144} & \textbf{282} \\
        \hline
            \parbox[t]{2mm}{\multirow{6}{*}{\rotatebox[origin=c]{90}{ResNet50}}} & SSD300-Resnet50 & 88 & 108 \\
            & SSD300 DCT LC-RFA & 87 & 110 \\
            & SSD300 DCT LC-RFA Y & 91 & 109 \\
            & SSD300 DCT LC-RFA-Thinner & 98 & 176 \\
            & SSD300 DCT LC-RFA-Thinner Y & 101 & 174 \\
            & SSD300 DCT Deconvolution-RFA & 87 & 104 \\
        \hline
        \end{tabular}
    \end{adjustbox}
    \label{table:ssd-speed}
\end{table}

\begin{table*}[!ht]
    \caption{Detection results on MS-COCO test-dev set. In bold are the best results of our trainings.}
    \centering
    \begin{adjustbox}{max width=0.8\linewidth}
        \begin{tabular}{l l | c c c | c c c | c c c | c c c}
        \hline
             & \multirow{2}{*}{Network} & \multicolumn{3}{c|}{Avg. Precision, IoU:} & \multicolumn{3}{c|}{Avg. Precision, Area:} & \multicolumn{3}{c|}{Avg. Recall, \#Dets:} & \multicolumn{3}{c}{Avg. Recall, Area:} \\
             & & 0.5:0.95 & 0.5 & 0.75 & S & M & L & 1 & 10 & 100 & S & M & L \\
        \hline
            & SSD300 \cite{ssd} & 23.2 & 41.2 & 23.4 & 5.3 & 23.2 & 39.6 & 22.5 & 33.2 & 35.3 & 9.6 & 37.6 & 56.5 \\
        \hline   
        \hline
            \parbox[t]{2mm}{\multirow{4}{*}{\rotatebox[origin=c]{90}{VGG}}} & SSD300 (our training) & 24.5 & 42.4 & 25.2 & \textbf{7.8} & 25.3 & 38.0 & 23.0 & 34.0 & 35.7 & \textbf{12.3} & 38.1 & 54.4 \\
            & SSD300 DCT & 14.3 & 27.0  & 13.7 & 2.1 & 12.1 & 26.2 & 15.8 & 22.4 & 23.4 & 3.4 & 21.1 & 42.5 \\
            & SSD300 DCT Y & 14.4 & 27.0 & 14.0 & 2.1 & 12.0 & 26.5 & 15.8 & 22.2 & 23.3 & 3.5 & 20.8 & 42.3 \\
            & SSD300 DCT Deconvolution & 13.5 & 26.0 & 12.6 & 2.5 & 11.3 & 23.8 & 15.3 & 21.9 & 23.1 & 4.5 & 21.1 & 39.6 \\
        \hline
            \parbox[t]{2mm}{\multirow{6}{*}{\rotatebox[origin=c]{90}{ResNet50}}} & SSD300 Resnet50 & \textbf{26.8} & \textbf{43.8} & \textbf{28.3} & 6.2 & \textbf{28.2} & \textbf{45.2} & \textbf{24.6} & \textbf{35.6} & \textbf{37.1} & 10.0 & \textbf{40.4} & \textbf{60.0} \\
            & SSD300 DCT LC-RFA & 25.8 & 42.4 & 27.1 & 5.1 & 27.0 & 44.4 & 23.9 & 34.2 & 35.6 & 8.0 & 38.8 & 59.0 \\
            & SSD300 DCT LC-RFA-Y & 25.2 & 41.6 & 26.5 & 5.2 & 25.7 & 43.7 & 23.6 & 33.7 & 35.0 & 8.1 & 37.4 & 58.2 \\
            & SSD300 DCT LC-RFA-Thinner & 25.4 & 41.8 & 26.9 & 4.7 & 26.3 & 44.6 & 23.7 & 33.8 & 35.1 & 7.2 & 38.0 & 59.4 \\
            & SSD300 DCT LC-RFA-Thinner-Y & 24.6 & 40.6 & 25.8 & 4.7 & 24.8 & 43.4 & 23.1 & 32.8 & 34.1 & 7.2 & 36.3 & 57.6 \\
            & SSD300 DCT Deconvolution-RFA & 25.9 & 42.5 & 27.2 & 5.4 & 26.7 & 44.4 & 24.0 & 34.5 & 36.0 & 8.5 & 39.0 & 59.4 \\
        \hline
        \end{tabular}
    \end{adjustbox}
    \label{table:ssd-coco}
\end{table*}

\paragraph{Comparison of the two RGB backbones}

Except when training on the \pascal{} 2007 set, we manage to reproduce the detection results for the SSD300 architecture. Regarding the \resnet{} based architecture, it performs worse than original SSD300 on the \pascal{} dataset but performs better on the \coco{} dataset. These results seem to indicate that the \resnet{} based architecture has a better convergence when provided with enough data. For both of the networks, FPS are mostly comparable.

\paragraph{Training with $YC_bC_r$ DCT inputs}

We detail all the architectures using the full compressed inputs, namely SSD300 DCT, SSD300 DCT LC-RFA and SSD300 DCT LC-RFA-Thinner. Deconvolution approaches are treated in the next section. For SSD300 DCT, we largely improve the results when compared with \cite{dct_detection}. We relate these improvements to a better trained backbone and to the incorporation of the dense layers as pre-trained convolutional layers. On the \pascal{} 07+12 dataset we reach 60.0 mAP and are 14.0 points behind the RGB method (18.9\% decrease). On the \coco{} dataset, we lose 10.1 points when compared with the RGB method but this time it represents a 42.2\% decrease. As could be expected, the hard pruning of the network is a limitation factor on more complex dataset such as \coco{}.

Regarding the \resnet{} based methods, SSD300 DCT LC-RFA and SSD300 DCT LC-RFA-Thinner outperform the VGG-based network by a large margin on the two datasets.
They even outperform the original SSD on \coco{} dataset. 
While this might be expected as the classification backbones used provide similar accuracy performances, the gap is sharper for detection. When comparing the SSD300 DCT LC-RFA with the SSD300 DCT LC-RFA-Thinner, we see that the second approach is 3 points behind the first on the \pascal{} datasets and about the same accuracy on \coco{}. The main advantage of the SSD300 DCT LC-RFA-Thinner is the number of FPS it can process, $\times 1.63$ more images than the SSD300 DCT LC-RFA while maintaining an equivalent accuracy.

\paragraph{Influence of the Deconvolution on detection performance}

While the deconvolution networks tend to perform better for classification we can see more mitigated results for detection. For the networks trained on the Pascal 2007 data only, we have a mAP of 38.4  for the VGG based network and 54.7 for the ResNet50 based one. They are respectively 12.3 points and 7.4 points lower than the mAP of the best performing DCT networks at equivalent speed (same backbone type, i.e VGG or ResNet). While the gap is reduced for the network trained on the 07+12 data, they still lag behind. However for the \coco{} dataset the SSD300 Deconvolution-RFA is the best performing of all the DCT based architectures. Moreover, when looking at results by size of area (Small, Medium or Large), we can see that both of the deconvolution architectures improve the accuracy for small objects. Overall, it seems that when provided with a enough data, the network will reach accuracy level equivalent to the non-deconvolution networks.

The SSD300 DCT Deconvolution is the fastest of all the detectors with a speed of  282 FPS but with the worst overall accuracy. The SSD300 Deconvolution-RFA as a speed equivalent to the RGB networks while not performing better than the SSD300 Resnet50.

\paragraph{Evaluation of the detection networks using only the $Y$ input}

On the \pascal{} dataset, we get similar mAP in comparison with the networks using the full $Y C_b C_r$ input. The reverse holds for the \coco{} dataset, where performances using only the Y input tend to be lower than their full input counter-parts (1 point below). Yet, it appears that the $C_b$ and $C_r$ components are not critical to correctly detect objects within images. Moreover,  networks for detection using only the $Y$ component have equivalent or higher speed than the networks using the $YC_bC_r$ inputs and require less bandwidth. The SSD300 DCT LC-RFA-Thinner Y network is $\times 1.70$ faster (with 2 instances, $\times 1.15$ when using only one instance of the network) than the original SSD while being only 3.8 point less accurate on the \pascal{} dataset and more accurate on the \coco{} dataset. The SSD300 DCT Y is even faster with a $\times 2.72$ speed improvement (with 2 instances, $\times 1.59$ when using only one instance) but at the cost of 14.2 points drop in the mAP for the \pascal{} dataset and 10.1 points for \coco{} dataset.

\paragraph{On the speed of the networks}

The FPS ratio the networks can process were computed either with one instance or with two instances of the model on one GPU. While this may seem anecdotal, we see that all the DCT network scale effortlessly when using 2 instances. The results for the network using compressed inputs indicates that they could be deployed on GPUs half as powerful as the one we used for the experiments and maintain the FPS obtained during the tests with 1 instance of the networks. This is not true for the RGB based network, which were already almost using all the computation capabilities of the GPU with only one model instantiated and thus would face important loss in regard to the FPS if deployed on GPUs half as powerful. This means that the presented architectures using the compressed inputs are good matches for usage in limited resources environment or on small remote computation devices. By combining our approach with other methods such as MobileNet \cite{mobilenet} or TinySSD \cite{tinyssd} we expect to even better fit such conditions.

\section{Conclusion}
\label{conclusion}

In this paper, we investigate object detection in \jpeg{} compressed images. We devise several deep architectures based on the SSD detection network framework. The architectures we explore differ in the classification backbones they rely on. Experimental evaluations evidence that they are not all equal for detection performance. 
%
When using \resnet{} as backbone we witness a slight performance drop of  5.5\% and 5.3\% on the \pascal{} and \coco{} datasets respectively in comparison to the best RGB network.
Furthermore, we demonstrate an effective speed up of $\times 1.7$ when using compressed input.
Also, we demonstrate that
using only the $Y$ input  
leads to detection performances similar to those of 
networks using the $YC_bC_r$ input. The benefit
is the reduced bandwidth for image transfer.
These findings are promising and may prove useful for the deployment of large real-time monitoring applications. 

\section*{Acknowledgment}

This  research  is  supported  by  ANRT  and  ACTEMIUM Paris Transport. We thank ACTEMIUM Paris Transport the funding. We thank CRIANN for the GPU computation facilities.



%

\bibliographystyle{unsrt}
\bibliography{bibliography}

\end{document}